\documentclass{article}

\usepackage{spconf,amsmath,graphicx}

\title{BiSNN: Training Spiking Neural Networks with Binary Weights \\ via Bayesian Learning}
\name{Hyeryung~Jang, Nicolas~Skatchkovsky, and Osvaldo~Simeone
\thanks{This work was supported by the European Research Council (ERC) under the European Union's Horizon 2020 research and innovation programme (grant agreement No. 725731) and by Intel Labs via the Intel Neuromorphic Research Community.
}
}
\address{KCLIP lab, CTR, Department of Engineering, King's College London, London, UK}
\usepackage[utf8]{inputenc}
\usepackage{microtype}
\usepackage{graphicx}
\usepackage{booktabs} % for professional tables
\usepackage{amsmath}
\usepackage{amsfonts}
\usepackage{amssymb}
\usepackage{amsthm}
\usepackage{pifont}
\usepackage{color}
\usepackage{array}
\usepackage{cite}
\usepackage{enumerate}
\usepackage{graphicx}
\usepackage[hang]{subfigure}
\usepackage{epstopdf}
\usepackage{epsfig}
\usepackage{footmisc}
\usepackage{amsbsy}
\usepackage{bm}
\usepackage{cases}
\usepackage{multirow}
\usepackage{algorithmic}
\usepackage{comment}
\usepackage{times}
\usepackage{paralist}
\usepackage[ruled]{algorithm2e}
\usepackage{setspace}
\usepackage[breaklinks]{hyperref}
\usepackage{mathtools}
\ninept

\newcommand{\bmx}{{\bm x}}
\newcommand{\bmy}{{\bm y}}

\newcommand{\bms}{{\bm s}}

\newcommand{\bmr}{{\bm r}}

\newcommand{\bmw}{{\bm w}}
\newcommand{\bmf}{{\bm f}}
\newcommand{\bml}{{\bm l}}

\newcommand{\bfp}{\mathbf{p}}

\newcommand{\bmepsilon}{\bm{\epsilon}}
\newcommand{\bmmu}{\bm{\mu}}
\newcommand{\bmdelta}{\bm{\delta}}

\newcommand{\set}[1]{\ensuremath{\mathcal #1}}
\newcommand{\grad}[1]{\nabla #1}

\begin{document}
%\ninept
%\bstctlcite{IEEEexample:BSTcontrol}
%
\maketitle
%
%% should remove this for submission
%\thispagestyle{plain}
%\pagestyle{plain}

%
\begin{abstract}
Artificial Neural Network (ANN)-based inference on battery-powered devices can be made more energy-efficient by restricting the synaptic weights to be binary, hence eliminating the need to perform multiplications. An alternative, emerging, approach relies on the use of Spiking Neural Networks (SNNs), biologically inspired, dynamic, event-driven models that enhance energy efficiency via the use of binary, sparse, activations. In this paper, an SNN model is introduced that combines the benefits of temporally sparse binary activations and of binary weights. Two learning rules are derived, the first based on the combination of straight-through and surrogate gradient techniques, and the second based on a Bayesian paradigm. Experiments validate the performance loss with respect to full-precision implementations, and demonstrate the advantage of the Bayesian paradigm in terms of accuracy and calibration. 
\end{abstract}
\begin{keywords}
Bayesian learning, Spiking Neural Networks, calibration, binary weights.
\end{keywords}

\section{Introduction}
\label{sec:intro}

%Much of the remarkable success in machine learning has relied on parametric models based on Artificial Neural Networks (ANNs) but it has been widely reported that ANNs often yield learning inference algorithms with massive energy, memory, and time requirements \cite{hao2019training}. The recent growth of edge devices, such as mobile phones, wearables, and Internet of Things (IoT) devices, has fueled the importance and need for low-power machine learning solutions on on-device edge intelligence. 
The deployment of machine learning models based on Artificial Neural Networks (ANNs) on mobile, battery-powered, devices enables a large number of applications, from personalized healthcare to environmental monitoring. Key performance metrics for mobile deployments are the energy and memory requirements that guarantee given accuracy levels. These generally depend on the type and amount of operations needed to process the input to produce the output prediction or action. 

A popular approach to enhance the energy efficiency of ANNs is to restrict the synaptic weights to binary $\{+1,-1\}$ values \cite{courbariaux2016binarized, rastegari2016xnor}. This makes it possible to avoid the use of expensive multiplications. A parallel, emerging, line of work, is investigating the use of Spiking Neural Networks (SNNs), a novel computational paradigm mimicking biological brains that is based on dynamic, event-driven, recursive processing of binary neural activations \cite{mead1990neuromorphic,snnreview}. By leveraging the temporal sparsity of binary time series signals, energy efficiency gains have been demonstrated on dedicated hardware implementations, such as IBM's TrueNorth \cite{akopyan2015truenorth} and Intel's Loihi \cite{davies2018loihi} (see also \cite{rajendran2019low}).

In this paper, we propose a model that aims at combining the efficiency gains of binary-weight neural implementations and of SNNs by introducing an SNN with binary $\{+1,-1\}$ synaptic weights. The resulting binary SNN model, referred to as BiSNN, is not only able to leverage temporal sparsity, through the use of a spiking neurons, but it can also reduce the complexity of neural operations via the use of binary weights. The adoption of binary weights has the added advantage of simplifying implementations based on beyond-CMOS memristive hardware \cite{adnan}. We specifically focus on the technical challenge of deriving learning algorithms for binary SNNs. Next, we review key issues and existing solutions.

Training binary ANNs is challenging because it involves optimizing over a large discrete space, making conventional continuous optimization methods inapplicable. The most common solution is based on the Straight-Through (ST) estimator \cite{bengio2013ste}. ST leverages latent, real-valued full-precision weights for the computation of the gradients, while relying on binarized weights for the forward pass through the network. The recent work \cite{meng2020bayesbinn} presents an alternative, theoretically principled, Bayesian framework that optimizes directly over the (continuous) distribution of the binary weights. The optimization problem is made tractable by following a mean-field variational inference approximation based on natural gradient descent \cite{khan2017conjugate}.

%based  Instead of optimizing a discrete binary weights, the Bayesian approach adopts a Bernoulli approximation to the posterior on the binary weights and optimizes it using natural gradient based variational inference method . This Bayesian approach enables to provide uncertainty estimates, allowing to combat a major shortcoming in ANNs. 

Training (full-precision) SNN models is also a complex problem owing to the non-differentiability of the threshold crossing-triggered binary activation of spiking neurons, whose gradient is zero almost everywhere. Most existing solutions view SNNs as Recurrent Neural Networks (RNNs), and adapt standard learning algorithms such as backpropagation through time (BPTT) via surrogate gradient (SG) techniques that smooth out the activation function \cite{huh2018gradient}. Alternatively, one can rely on probabilistic models for spiking neurons \cite{jang19:spm, jimenez2014stochastic, brea2013matching} or on conversion from a pre-trained ANN \cite{rueckauer2018conversion} (see \cite{snnreview} for a review).

In this work, we first propose to combine ST and SG techniques to define a frequentist training algorithm for BiSNNs, termed ST-BiSNN. Then, we introduce a novel learning algorithm -- Bayes-BiSNN -- that follows generalized Bayesian principles, or Information Risk Minimization \cite{zhang2006itbound}. We note that the recent work \cite{lu2020exploring} has also explored BiSNNs, but did not propose any direct training algorithm, relying instead on conversion methods from ANNs. Such conversion methods are known to be generally inefficient as compared to the direct training of SNN models \cite{rosenfeld2019learning,snnreview}. Through experiments on both synthetic and real neuromorphic data sets, we validate the performance loss with respect to full-precision implementations, and demonstrate the advantage of the Bayesian paradigm in terms of accuracy and calibration.

%The remainder of this paper is organized as follows. In Sec.~\ref{sec:model}, we review the details of the binary SNN models. In Sec.~\ref{sec:st-bisnn} and Sec.~\ref{sec:bayes-bisnn}, we introduce the ST-BiSNN and Bayes-BiSNN training rules for binary SNNs. In Sec~\ref{sec:exp}, we demonstrate the strengths and weaknesses of the proposed schemes, and compare them with standard rules for real-valued weights. 

%The focus in the training of traditional ANNs has recently shifted from traditional algorithms such as backpropagation to principled techniques based on a \textit{bayesian} perspective to capture uncertainty through the estimation of posterior distributions on the model parameters. This allows to combat a major shortcoming in ANNs, that is the overconfidence in the networks' predictions \note{refs}, especially is the face of unseen data, that severely limits the practicability of ANNs for critical applications such as medicine or self-driving cars. However, exact posterior inference is intractable in ANNs \note{refs}, and most existing bayesian techniques rely on crude approximations that limit the quality of inference \note{refs}. 

\section{Model and Problem}
\label{sec:model}

In this paper, we design training algorithms for binary SNNs (BiSNNs) in which, unlike conventional SNN models, the synaptic weights can only take binary values $\{+1,-1\}$. 
An BiSNN is defined by a network $\set{V}$ of spiking neurons connected over an arbitrary graph, which possibly includes (directed) cycles. 
Focusing on a discrete-time implementation, each spiking neuron $i \in \set{V}$ produces a binary value $s_{i,t} \in \{0,1\}$ at discrete time $t=1,2,\ldots$, with ``$1$'' denoting the firing of a spike. 
We collect in $|\set{V}| \times 1$ vector $\bms_t = (s_{i,t}: i \in \set{V})$ the spikes emitted by all neuron $\set{V}$ at time $t$, and denote by $\bms_{\leq t} = (\bms_1, \ldots, \bms_t)$ the spike sequences of all neurons up to time $t$. 
Each neuron $i$ receives input spike signals $\{s_{j,t}\}_{j \in \set{P}_i} = \bms_{\set{P}_i, t}$ from the set $\set{P}_i$ of parent, or pre-synaptic, neurons, which is connected to neuron $i$ via directed links in the graph. With some abuse of notations, this set is taken to include also exogeneous input signals.

Each neuron $i$ maintains a scalar analog state variable $u_{i,t}$ known as membrane potential, and it outputs the binary signal 
\begin{align} \label{eq:ind-threshold}
    s_{i,t} = \Theta(u_{i,t} - \vartheta),
\end{align}
with $\Theta(\cdot)$ being the Heaviside step function and $\vartheta$ being a fixed firing threshold. 
Accordingly, a spike is produced when the membrane potential is above threshold $\vartheta$. 
Following the standard discrete-time Spike Response Model (SRM), the membrane potential $u_{i,t}$ is obtained by summing filtered contributions from pre-synaptic neurons in set $\set{P}_i$ and from the neuron's own output. In particular, the membrane potential evolves as
\begin{align} \label{eq:ind-membrane}
    u_{i,t} = \sum_{j \in \set{P}_i} w_{ij} \big( \underbrace{{\alpha_t} \ast s_{j,t}}_{\triangleq~ p_{j,t}} \big) - \beta_t \ast s_{i,t},
\end{align}
where $w_{ij} \in \{+1,-1\}$ is a learnable binary synaptic weight from pre-synaptic neuron $j \in \set{P}_i$; $\alpha_t$ and $\beta_t$ represent the spike responses of synapses and somas, respectively; and $\ast$ denotes the convolution operator $f_t \ast g_t = \sum_{\delta > 0} f_{\delta} g_{t-\delta}$. 
Typical choices are the second-order synaptic filter $\alpha_t = \exp(-t/\tau_{\text{mem}}) - \exp(-t/\tau_{\text{syn}})$, known as $\alpha$-function \cite{gerstner2002spiking}, and the first-order feedback filter $\beta_t = \exp(-t/\tau_{\text{ref}})$, for $t \geq 1$ and finite positive constants $\tau_{\text{mem}}, \tau_{\text{syn}}$, and $\tau_{\text{ref}}$. 
With these filters, the updates in \eqref{eq:ind-membrane} can be implemented using recursive equations \cite{kaiser2020decolle}.

In BiSNNs, the synaptic weight $w_{ij}$ are binary, i.e., $w_{ij} \in \{+1,-1\}$. 
This ensures that no products are needed to compute the membrane potential $u_{i,t}$ in \eqref{eq:ind-membrane}, hence significantly reducing complexity. 
Furthermore, BiSNNs are particularly well suited for hardware implementations on chips with nanoscale components that provide discrete conductance levels for the synapses \cite{adnan}. In this regard, we note that the results in this paper can be generalized to weights with any number of finite values.

We divide the neurons of the BiSNN into a read-out, or output, layer $\set{Y}$ and a set of hidden neurons $\set{H}$, with $\set{V} = \set{H} \cup \set{Y}$. The set of exogeneous inputs is defined as $\set{X}$. In supervised learning, a data set $\set{D}$ is given by $N$ pairs $(\bmx_{\leq T}, \bmr_{\leq T})$ of signals generated from an unknown distribution $p(\bmx_{\leq T}, \bmr_{\leq T})$, with $\bmx_{\leq T}$ being a vector of signals up to time $T$ from the set $\set{X}$ of exogeneous inputs and $\bmr_{\leq T}$ the desired reference signals for the set $\set{Y}$ of visible output neurons. We note that the exogeneous inputs and target reference signals need not to be binary -- or spiking -- although this is the case for many applications of interest, which allows us to include implementations with per-layer local surrogate losses \cite{kaiser2020decolle}.

We write as $\bmf_{\bmw}(\bmx_{\leq T})$ the spiking signals emitted by the set $\set{Y}$ of neurons in the output layer, following the SRM \eqref{eq:ind-threshold}-\eqref{eq:ind-membrane} in response to exogeneous inputs $\bmx_{\leq T}$, where $\bmw := \{\{w_{ij}\}_{j \in \set{P}_i}\}_{i \in \set{V}}$ collects the model parameters. 
We also write $( \bmf_{\bmw}(\bmx_{\leq t}))_t$ for the $|\set{Y}| \times 1$ vector of output signals produced by the set $\set{Y}$ of neurons at time $t$, and $( \bmf_{\bmw}(\bmx_{\leq t}))_{i,t}$ as the output signal of output neuron $i \in \set{Y}$ at time $t$.

The goal is to minimize the average loss entailed by the output signals $\bmf_{\bmw}(\bmx_{\leq T})$ produced by the SNN. The loss is measured with respect to a reference signal $\bmr_{\leq T}$. The reference signal $\bmr_{\leq T} = (\bmr_1,\ldots, \bmr_T)$ may directly represent the desired output spiking signals of neurons in set $\set{Y}$; or it may define more general supervisory signals, such as the class index corresponding to the current input $\bmx_{\leq T}$ \cite{kaiser2020decolle}. 
The corresponding population loss is given as 
\begin{align} \label{eq:gen-loss}
    \set{L}_{p}(\bmw) = \mathbb{E}_{ p(\bmx_{\leq T}, \bmr_{\leq T})} \Big[ \ell \big( \bmf_{\bmw}(\bmx_{\leq T}), \bmr_{\leq T} \big) \Big],
\end{align}
where $\ell(\bmy_{\leq T}, \bmr_{\leq T})$ measures the loss produced by spiking signals $\bmy_{\leq T}$ given reference signals $\bmr_{\leq T}$ and the average is over the unknown population distribution $p(\bmx_{\leq T}, \bmr_{\leq T})$. We assume that the loss in \eqref{eq:gen-loss} can be written as\begin{align} \label{eq:loss-real}
    \ell\big( \bmf_{\bmw}(\bmx_{\leq T}), \bmr_{\leq T} \big) = \sum_{t=1}^T \bml_t(\bmw),
\end{align} where $\bml_t(\bmw)$ is the loss due to the actual output signal $\bmy_t = ( \bmf_{\bmw}(\bmx_{\leq t}))_t$ given the reference signal $\bmr_t$ at time $t$. As an example, in \cite{kaiser2020decolle}, the loss $\bml_t(\bmw)$ is chosen as the logistic loss between the output of the softmax function of a fixed, random classification layer with inputs given by $\bmy_t$ and the class label $\bmr_t$.

Since the population distribution $p(\bmx_{\leq T}, \bmr_{\leq T})$ is unknown, training uses as objective the empirical estimate of the population loss $\set{L}_p(\bmw)$ in \eqref{eq:gen-loss} based on the $N$ data samples in the training data set $\set{D} = \{(\bmx_{\leq T}, \bmr_{\leq T})\}$, and is formulated as the problem 
\begin{align} \label{eq:loss}
    \min_{\bmw \in \{+1,-1\}^{|\bmw|}}~ \set{L}_{\set{D}}(\bmw) ~ = \frac{1}{N} ~~  \smashoperator{\sum_{(\bmx_{\leq T}, \bmr_{\leq T}) \in \set{D}}} ~~ \ell\big( \bmf_{\bmw}(\bmx_{\leq T}), \bmr_{\leq T} \big).
\end{align}
Problem \eqref{eq:loss} cannot be solved using standard gradient-based methods since: {\em (i)} function $\bmf_{\bmw}(\cdot)$ is not differentiable in $\bmw$ due to the presence of the threshold function \eqref{eq:ind-threshold}; and {\em (ii)} the domain of the weight vector $\bmw$ is the discrete set of binary values. In the next section, we introduce a solution based on the ST gradient estimator \cite{bengio2013ste} and SG techniques \cite{neftci2019surrogate} that directly tackles problem \eqref{eq:loss} through smooth approximations of the training loss $\set{L}_{\set{D}}(\bmw)$. Then, we propose a conceptually different approach based on Bayesian principles and Information Risk Minimization (IRM) \cite{zhang2006itbound}.

\section{ST-BiSNN: Straight-Through Learning for BiSNNs} 
\label{sec:st-bisnn}

In this section, we review a solution that tackles problem \eqref{eq:loss} by combining surrogate gradient techniques \cite{neftci2019surrogate} and the ST estimator \cite{bengio2013ste} to address non-differentiability and the binary constraints, respectively. In the resulting ST-BiSNN rule, the gradient is computed using surrogate gradient methods with respect to ``latent'' real-valued weights $\bmw^\text{r}$, which are quantized to obtain the next binary iterate $\bmw$. The algorithm keeps track of the real-vector $\bmw^\text{r}$ in order to avoid the accumulation of quantization noise across the iterations. The surrogate gradient method approximates the Heaviside step function $\Theta(\cdot)$ in \eqref{eq:ind-threshold} with a differentiable sigmoid function $\sigma(\cdot)$ in order to obtain non-zero gradients of the training loss $\set{L}_{\set{D}}(\bmw)$ in \eqref{eq:loss}.

To elaborate, we define as $\bmw^\text{r} \in \mathbb{R}^{|\bmw|}$ a relaxation of $\bmw$ to the real vector space. 
The partial derivative of the loss $\ell$ in \eqref{eq:loss} with respect to each real-valued synaptic weight $w_{ij}^\text{r}$ is obtained as
\begin{align} \label{eq:partial-der}
    \frac{\partial \ell\big( \bmf_{\bmw^{\text{r}}}(\bmx_{\leq T}), \bmr_{\leq T}\big)}{\partial w_{ij}^{\text{r}}} 
    = \sum_{t=1}^T \underbrace{ \frac{\partial \bml_t(\bmw^\text{r})}{\partial \big(\bmf_{\bmw^\text{r}}(\bmx_{\leq t})\big)_t} \frac{\partial \big(\bmf_{\bmw^\text{r}}(\bmx_{\leq t})\big)_t}{\partial s_{i,t}} }_{:=~ e_{i,t}:~ \text{error signal}}  \frac{s_{i,t}}{\partial w_{ij}^{\text{r}}},
\end{align}
where the first term can be interpreted as a (scalar) error signal $e_{i,t}$ for the post-synaptic neuron $i$, and the second term is the derivative of the post-synaptic neuron's output.

\DecMargin{2em}
\begin{algorithm}[t]
\caption{ST-BiSNN}
\label{alg:st-bisnn}
\begin{algorithmic}[1]
   \STATE {\bfseries Input:} data set $\set{D}$, learning rate $\eta$
   
   \STATE{\bfseries Output:} learned binary weights $\bmw$
   
   \vspace{-0.2cm}
   \hrulefill
   \STATE {\bf initialize} real-valued weights $\bmw^\text{r}$
   
   \REPEAT
   \STATE select mini-batch $\set{B} = \{(\bmx_{\leq T}, \bmr_{\leq T})\} \subseteq \set{D}$
   \smallskip
   \STATE quantize $\bmw^\text{r}$ to obtain binary weights $\bmw$ as
   \begin{align} \label{eq:st-binarize}
       \bmw = \text{sign}\big( \bmw^\text{r}\big)
   \end{align}
   
   %\smallskip
   \STATE for each $(\bmx_{\leq T}, \bmr_{\leq T}) \in \set{B}$, compute the gradient  $\grad_{\bmw^\text{r}} \ell \big( \bmf_{\bmw^\text{r}}(\bmx_{\leq T}), \bmr_{\leq T}\big) \big|_{\bmw^\text{r} = \bmw}$ using \eqref{eq:der-general}
   
   \medskip
   \STATE update the real-valued weights $\bmw^\text{r}$ via SGD as
   \begin{eqnarray*} %\label{eq:st-update}
       \bmw^\text{r} \leftarrow \bmw^\text{r} - \frac{\eta}{|\set{B}|} \cdot ~~~ \smashoperator{\sum_{(\bmx_{\leq T}, \bmr_{\leq T}) \in \set{B}}} ~ \grad_{\bmw^\text{r}} \ell \big( \bmf_{\bmw^\text{r}}(\bmx_{\leq T}), \bmr_{\leq T}\big) \Big|_{\bmw^\text{r} = \bmw}
   \end{eqnarray*}
%   \begin{eqnarray} \label{eq:st-update}
%       \bmw^\text{r} \leftarrow \bmw^\text{r} - \frac{\eta}{|\set{B}|} \cdot ~ \smashoperator{\sum_{(\bmx_{\leq T}, \bmy_{\leq T}) \in \set{B}}} ~ \grad_{\bmw^\text{r}} \ell \big( \bmf_{\bmw^\text{r}}(\bmx_{\leq T}), \bmy_{\leq T}\big) \Big|_{\bmw^\text{r} = \bmw}
%   \end{eqnarray}
   \UNTIL{convergence}
\end{algorithmic}
\end{algorithm}
\IncMargin{2em}

From \eqref{eq:ind-threshold}-\eqref{eq:ind-membrane}, the term $\partial s_{i,t} / \partial w_{ij}^\text{r}$ depends on the derivative of the Heaviside step function $\Theta(u_{i,t}-\vartheta)$, which is zero almost everywhere. Following \cite{neftci2019surrogate}, we adopt the surrogate gradient method by replacing the derivative $\Theta'(\cdot)$ with the derivative $\sigma'(\cdot) = \sigma(\cdot)(1-\sigma(\cdot))$ of a sigmoid function $\sigma(\cdot)$, obtaining the approximation 
\begin{align}
    \frac{\partial s_{i,t}}{\partial w_{ij}^{\text{r}}} %= \frac{\partial \Theta(u_{i,t} - \vartheta)}{\partial w_{ij}^{\text{r}}} 
    = \Theta'(u_{i,t} - \vartheta) \frac{\partial u_{i,t}}{\partial w_{ij}^\text{r}} \approx \sigma'(u_{i,t}-\vartheta) p_{j,t},
\end{align}
where we have $\frac{\partial u_{i,t}}{\partial w_{ij}^\text{r}} = p_{j,t}$ from \eqref{eq:ind-membrane}. The resulting derivative in \eqref{eq:partial-der} is then given as 
\begin{align} \label{eq:der-general}
    \frac{\partial \ell\big( \bmf_{\bmw^{\text{r}}}(\bmx_{\leq T}), \bmr_{\leq T}\big) }{\partial w_{ij}^{\text{r}}} = \sum_{t=1}^T \underbrace{ e_{i,t} }_{\text{error signal}} \underbrace{ \sigma'(u_{i,t}-\vartheta) }_{\text{post}_i} \underbrace{ p_{j,t} }_{\text{pre}_j},
\end{align}
% \begin{multline} \label{eq:der-general}
%     \frac{\partial \ell\big( \bmf_{\bmw^{\text{r}}}(\bmx_{\leq T}), \bmr_{\leq T}\big) }{\partial w_{ij}^{\text{r}}} \\
%     = \sum_{t=1}^T \underbrace{ e_{i,t} }_{\text{error signal}} \cdot \Big( \underbrace{ \sigma'(u_{i,t}-\vartheta) }_{\text{post}_i} \underbrace{ p_{j,t} }_{\text{pre}_j}\Big),
%     % = \sum_{t=1}^T \underbrace{ e_{i,t} }_{\text{error signal}} \Big( \gamma_t \ast \big( \underbrace{ \sigma'(u_{i,t}-\vartheta) }_{\text{post}_i} \underbrace{ p_{j,t} }_{\text{pre}_j} \big) \Big),
%     %& \quad = \sum_{t=1}^T \Big( \underbrace{ \gamma_t \ast \Big( \bml_t(\bmw^{\text{r}})^{\top} \frac{ \partial \bml_t(\bmw^{\text{r}})}{\partial s_{i,t}} \Big) }_{ := e_{i,t}: ~\text{error signal}} \Big) \cdot \cr
%     %& \qquad \Big( \gamma_t \ast \Big(\underbrace{ \sigma'(u_{i,t}-\vartheta) }_{\text{post}_i} \big( \underbrace{ p_{j,t} }_{\text{pre}_j} \big) \Big) \Big),
% \end{multline}
where we have highlighted the product of error signal, post-synaptic term $\sigma'(u_{i,t}-\vartheta)$ and pre-synaptic trace $p_{j,t}$. 
The error signal $e_{i,t}$ in \eqref{eq:partial-der} can be in principle computed via backpropagation through time. 
As in \cite{neftci2019surrogate, zenke2018superspike}, simpler feedback schemes that do not require signal propagation along the backward path can be implemented to obtain the error signals $e_{i,t}$. For instance, in \cite{kaiser2020decolle}, the error signal $e_{i,t}$ is obtained in terms of a local logistic loss with randomized projections of the per-layer spiking outputs.

ST-BiSNN proceeds iteratively by selecting a mini-batch $\set{B}$ of examples $(\bmx_{\leq T}, \bmr_{\leq T})$ from the training data set $\set{D}$ at each iteration. Binary and real-valued synaptic weight vectors $\bmw$ and $\bmw^{\text{r}}$ are updated as described in Algorithm~\ref{alg:st-bisnn}. The sign function is defined as $\text{sign}(x) = +1$ for $x \geq 0$ and $\text{sign}(x) = -1$ for $x < 0$ and is applied in \eqref{eq:st-binarize} element-wise for quantization.

\iffalse
As an alternative, we consider also the Binary optimizer (Bop) method  \cite{helwegen2019bop} as applied to SNN, referred to as Bop-BiSNN:

\begin{itemize}  
    \item For the current real-valued vector $\bmw^{\text{r}}$, obtain the binary weight vector $\bmw$ as 
    \begin{align}
        \bmw \leftarrow \text{hyst}\big(\bmw^{\text{r}}, \bmw\big),
    \end{align}
    where the hysteresis function $\text{hyst}(\cdot)$ is defined element-wise as
    \begin{align}
        \text{hyst}\big(w_{ij}^{\text{r}}, w_{ij}\big) = \begin{cases} 
        -w_{ij}, ~~\text{if}~ |w_{ij}^{\text{r}}| > \tau_{\text{hyst}} ~\text{and}~ \cr
        \quad \qquad \text{sign}(w_{ij}^{\text{r}}) = \text{sign}(w_{ij}), \\ 
        w_{ij}, ~~~~\text{otherwise},
        \end{cases}
    \end{align}
    for some threshold constant $\tau_{\text{hyst}} > 0$. 
    
    \item At the current binary iterate $\bmw$, compute the gradient $\grad_{\bmw^{\text{r}}} \ell\big( \bmf_{\bmw^{\text{r}}}(\bmx_{\leq T}), \bmy_{\leq T} \big)\bigr|_{\bmw^{\text{r}} = \bmw}$.
    
    \item Update the real-valued weight vector $\bmw^{\text{r}}$ via SGD as 
    \begin{align}
        \bmw^{\text{r}} \leftarrow (1-\eta) \bmw^{\text{r}} + \eta \grad_{{\bmw^{\text{r}}}} \ell\big( \bmf_{\bmw^{\text{r}}}(\bmx_{\leq T}), \bmy_{\leq T} \big)\bigr|_{\bmw^{\text{r}} = \bmw},
    \end{align}
    with learning rate $\eta > 0$.
    
\end{itemize}
\fi

\section{Bayes-BiSNN: Bayesian Learning for BiSNNs} 
\label{sec:bayes-bisnn}

%include only Bayes-BiSNN-RP

In this section, we introduce a new learning rule, Bayes-BiSNN, that tackles the problem of training BiSNNs within a generalized Bayesian framework. Accordingly, we formulate the training problem as the minimization over a probability distribution $q(\bmw)$ in the space of binary weights, which is referred to as variational posterior. Specifically, following the IRM, or generalized Bayesian, formulation \cite{zhang2006itbound}, we aim at solving the problem 
\begin{align} \label{eq:bayes-opt}
    \min_{q(\bmw)}~ \mathbb{E}_{q(\bmw)} \Big[ \set{L}_{\set{D}}(\bmw) \Big] + \rho \text{KL}\big( q(\bmw) || p(\bmw) \big),
\end{align}
where $\rho > 0$ is a temperature constant, $p(\bmw)$ is an arbitrary prior distribution over binary weights, and $\text{KL}(\cdot||\cdot)$ is the Kullback-Leibler divergence $\text{KL}(q(\bmw)||p(\bmw)) = \mathbb{E}_{q(\bmw)}[ \log (q(\bmw)/p(\bmw))]$. The objective function in \eqref{eq:bayes-opt} is known as free energy \cite{jose2020free}.

The problem \eqref{eq:bayes-opt} of minimizing the free energy must strike a balance between fitting the data -- i.e., minimizing the first term -- and not deviating too much from the reference behavior defined by prior $p(\bmw)$ -- i.e., keeping the second term small. The KL divergence term can be thought of as a regularizing penalty that accounts for epistemic uncertainty due to the presence of limited data \cite{zhang2006itbound} or for the complexity of information processing \cite{jose2020free}. If no constraints are imposed on the variational posterior $q(\bmw)$, the optimal solution of \eqref{eq:bayes-opt} is given by the Gibbs posterior
\begin{align} \label{eq:bayes-sol}
    q^\star(\bmw) = \frac{ p(\bmw) \exp\big( - \set{L}_{\set{D}}(\bmw) / \rho \big)}{\mathbb{E}_{p(\bmw)} \Big[ \exp\big( - \set{L}_{\set{D}}(\bmw) / \rho \big) \Big]}.
\end{align}
Due to intractability of the normalizing constant in \eqref{eq:bayes-sol}, we adopt instead a mean-field Bernoulli variational approximation by limiting the optimization domain for problem \eqref{eq:bayes-opt} to variational posteriors of the form $q_{\bfp}(\bmw) = \text{Bern}\big( \bmw|\bfp \big)$ as 
\begin{align} \label{eq:exp-bernoulli}
    q_{\bfp}(\bmw) = \prod_{i \in \set{V}} \prod_{j \in \set{P}_i} p_{ij}^{\frac{1+w_{ij}}{2}} (1-p_{ij})^{\frac{1-w_{ij}}{2}},
\end{align}
where $p_{ij}$ is the probability that synaptic weight $w_{ij}$ equals $+1$, and we have defined the vector $\bfp = \{\{p_{ij}\}_{j \in \set{P}_i}\}_{i \in \set{V}}$ to collect all variational parameters.

The variational posterior \eqref{eq:exp-bernoulli} can be reparameterized in terms of the mean parameters $\bmmu = \{\{\mu_{ij}\}_{j \in \set{P}_i}\}_{i \in \set{V}}$ as $q_{\bmmu}(\bmw) = \text{Bern}\big( \bmw | \frac{\bmmu+1}{2} \big)$ by setting $p_{ij} = (\mu_{ij}+1)/2$. It can also be expressed in terms of the logits, or natural parameters, $\bmw^\text{r} = \{\{w_{ij}^\text{r}\}_{j \in \set{P}_i}\}_{i \in \set{V}}$ as $q_{\bmw^\text{r}}(\bmw) = \text{Bern}\big( \bmw | \sigma(2\bmw^\text{r}) \big)$ by setting 
\begin{align} \label{eq:bernoulli-param}
    w_{ij}^{\text{r}} = \frac{1}{2} \log \bigg(\frac{p_{ij}}{1-p_{ij}}\bigg) = \frac{1}{2} \log \bigg( \frac{1+\mu_{ij}}{1-\mu_{ij}} \bigg).
\end{align}
As we will see, the notation $\bmw^\text{r}$ has been introduced in \eqref{eq:bernoulli-param} to suggest a relationship with the ST-BiSNN method described in Sec.~\ref{sec:st-bisnn}. We assume that the prior distribution $p(\bmw)$ also follows the mean-field Bernoulli distribution of the form $p(\bmw) = \text{Bern}(\bmw | \sigma(2\bmw^\text{r}_0))$, with $\bmw^\text{r}_0$ being the corresponding logits, thus $\bmw^\text{r}_0 = {\bm 0}$ if the binary weights are equally likely to be either $+1$ or $-1$ as a priori. 

\DecMargin{2em}
\begin{algorithm}[t]
\caption{Bayes-BiSNN}
\label{alg:bayes-bisnn}
\begin{algorithmic}[1]
   \STATE {\bfseries Input:} data set $\set{D}$, learning rate $\eta$, temperature parameter $\rho$, GS trick parameter $\tau$, logits $\bmw^\text{r}_0$ of prior distribution 
   
   \STATE{\bfseries Output:} learned binary weights $\bmw$
   
   \vspace{-0.2cm}
   \hrulefill
   
   \STATE {\bf initialize} real-valued weights $\bmw^\text{r}$
   
   \REPEAT
   \STATE select mini-batch $\set{B} = \{(\bmx_{\leq T}, \bmr_{\leq T})\} \subseteq \set{D}$
   \smallskip
   \STATE sample relaxed binary weights as
   \begin{align*}
       \bmw = \text{tanh}\bigg( \frac{\bmw^\text{r}+\bmdelta}{\tau}\bigg),
   \end{align*}
   with $\bmdelta = \frac{1}{2} \log \frac{\bmepsilon}{1-\bmepsilon}$ and $\bmepsilon \stackrel{\text{i.i.d.}}{\sim} \set{U}(0,1)$, with all operations being element-wise
   
   \smallskip
   \STATE for each $(\bmx_{\leq T}, \bmr_{\leq T}) \in \set{B}$, compute the gradient  $\grad_{\bmw^\text{r}} \ell \big( \bmf_{\bmw^\text{r}}(\bmx_{\leq T}), \bmr_{\leq T}\big) \big|_{\bmw^\text{r} = \bmw}$ using \eqref{eq:der-general}
   
   \smallskip
   \STATE compute unbiased gradient estimator $\grad_{\bmmu}\hat{\set{L}}(\bmmu)$ from \eqref{eq:bayes-RP-grad-estimator} and update the real-valued weights $\bmw^\text{r}$ as
   \begin{align*} %\label{eq:bayes-RP-update}
       \bmw^\text{r} \leftarrow (1-\eta \rho) \bmw^\text{r} - \eta \Big( \grad_{\bmmu} \hat{\set{L}}(\bmmu) - \rho \bmw^\text{r}_0 \Big)
   \end{align*}
%   \begin{align} \label{eq:bayes-RP-update}
%       \bmw^\text{r} \leftarrow (1-\eta) \bmw^\text{r} - \frac{\eta}{\rho} \grad_{\bmmu} \hat{\set{L}}(\bmmu)
%   \end{align}

   \UNTIL{convergence}
\end{algorithmic}
\end{algorithm}
\IncMargin{2em}

\begin{table*}[th!]
\caption{Comparison of ST-BiSNN and Bayes-BiSNN training rules. Bayes-BiSNN assumes a uniform prior distribution $p(\bmw)$.}
\label{tbl:bisnn-grad}
\vskip 0.15in
\begin{center}
\begin{small}
%\begin{sc}
\begin{tabular}{lcc}
\toprule
Scheme & Binarization & Update rule with $(\bmx_{\leq T}, \bmr_{\leq T}) \in \set{D}$ \\
\midrule
ST-BiSNN & $\bmw = \text{sign}\big( \bmw^\text{r} \big)$ & $\bmw^\text{r} \leftarrow \bmw^\text{r} - \eta \grad_{\bmw^{\text{r}}} \ell( \bmf_{\bmw^{\text{r}}}(\bmx_{\leq T}), \bmr_{\leq T}) \bigr|_{\bmw^{\text{r}} = \bmw}$ \\[4pt]
Bayes-BiSNN & $\bmw = \text{tanh}\Big( \frac{\bmw^{\text{r}} + \bmdelta}{\tau} \Big)$ & $\bmw^\text{r} \leftarrow (1-\eta \rho) \bmw^\text{r} - \eta \frac{1-\bmw^2}{\tau (1 - \text{tanh}^2(\bmw^{\text{r}}))} \odot \grad_{\bmw^{\text{r}}} \ell( \bmf_{\bmw^{\text{r}}}(\bmx_{\leq T}), \bmr_{\leq T}) \bigr|_{\bmw^{\text{r}} = \bmw}$ \\[4pt]
\bottomrule
\end{tabular}
%\end{sc}
\end{small}
\end{center}
\vspace{-0.5cm}
\end{table*}

Bayes-BiSNN applies the natural gradient rule to minimize the free energy \eqref{eq:bayes-opt} with respect to the variational parameters $\bmw^\text{r}$ defining the variational posterior $q_{\bmw^\text{r}}(\bmw)$. Following \cite{meng2020bayesbinn}, this yields the update 
\begin{align} \label{eq:bayes-update}
    \bmw^{\text{r}} \leftarrow (1-\eta \rho) \bmw^{\text{r}} - \eta \bigg( \grad_{\bmmu} \mathbb{E}_{q_{\bmw^\text{r}}(\bmw)} \Big[ \set{L}_{\set{D}}(\bmw) \Big] - \rho \bmw^\text{r}_0 \bigg),
\end{align}
where $0 < \eta < 1$ is the learning rate. Note that the gradient in \eqref{eq:bayes-update} is with respect to the mean parameters $\bmmu$. We also point to the paper \cite{kreutzer2020natural}, which explores the use of natural gradient descent for frequentist learning in spiking neurons. In order to estimate the gradient in \eqref{eq:bayes-update}, Bayes-BiSNN leverages the {\em reparameterization} trick via the {\em Gumbel-Softmax} (GS) distribution \cite{jang2016categorical, meng2020bayesbinn}. Accordingly, we first obtain one sample $\bmw$ that is approximately distributed according to $q_{\bmw^\text{r}}(\bmw)$ with $\bmw^\text{r}$ and $\bmmu$ related through \eqref{eq:bernoulli-param}. This is done by drawing a vector $\bmdelta = \{\{\delta_{ij}\}_{j \in \set{P}_i}\}_{i \in \set{V}}$ of i.i.d. Gumbel variables, which we denote as $\bmdelta \sim p(\bmdelta)$, and then computing 
\begin{align} \label{eq:GS-trick}
    \bmw = \text{tanh}\bigg( \frac{\bmw^{\text{r}} + \bmdelta}{\tau} \bigg),
\end{align}
where $\tau > 0$ is a parameter; and the $\text{tanh}(\cdot)$ function is applied element-wise. When $\tau$ in \eqref{eq:GS-trick} tends to zero, the $\text{tanh}(\cdot)$ function tends to the $\text{sign}(\cdot)$ function, and the vector $\bmw$ follows distribution $q_{\bmw^{\text{r}}}(\bmw)$ \cite{meng2020bayesbinn}. To generate $\bmdelta$, one can set $\delta_{ij} = \frac{1}{2} \log \left(\frac{\epsilon_{ij}}{1-\epsilon_{ij}}\right)$, with $\epsilon_{ij} \sim \set{U}(0,1)$ being i.i.d. samples.

With this sample, we then obtain an approximately unbiased estimate of the gradient in \eqref{eq:bayes-update} by using the following approximate equality 
\begin{multline} \label{eq:bayes-RP-grad}
    \grad_{\bmmu} \mathbb{E}_{q_{\bmw^\text{r}}(\bmw)} \Big[ \set{L}_{\set{D}}(\bmw) \Big] \stackrel{(a)}{\approx} \mathbb{E}_{p(\bmdelta)} \bigg[ \grad_{\bmmu} \set{L}_{\set{D}}\Big( \text{tanh}\Big(\frac{\bmw^\text{r}+\bmdelta}{\tau}\Big) \Big)\bigg] \cr 
    ~~ \stackrel{(b)}{=} \mathbb{E}_{p(\bmdelta)} \bigg[ \grad_{\bmw} \set{L}_{\set{D}}(\bmw) \odot \grad_{\bmmu} \text{tanh}\Big( \frac{\bmw^\text{r}+\bmdelta}{\tau} \Big) \bigg] \cr 
    = \mathbb{E}_{p(\bmdelta)} \bigg[ \grad_{\bmw} \set{L}_{\set{D}}(\bmw) \odot \frac{1-\bmw^2}{\tau \big( 1 - \text{tanh}^2(\bmw^\text{r}) \big)} \bigg],
\end{multline}
where the approximate equality (a) is exact when $\tau \rightarrow 0$ and the equality (b) follows the chain rule. In \eqref{eq:bayes-RP-grad}, the symbol $\odot$ denotes the element-wise product. We note that the gradient $\grad_{\bmw} \set{L}_{\set{D}}(\bmw)$ can be computed using the loss gradient $\grad_{\bmw^\text{r}} \ell ( \bmf_{\bmw^\text{r}}(\bmx_{\leq T}), \bmr_{\leq T}) |_{\bmw^\text{r} = \bmw}$ from \eqref{eq:der-general}.

As summarized in Algorithm~\ref{alg:bayes-bisnn}, the resulting Bayes-BiSNN rule proceeds iteratively by selecting a mini-batch $\set{B}$ of examples $(\bmx_{\leq T}, \bmr_{\leq T})$ from the training data set $\set{D}$ at each iteration. Using the sample $\bmw$ from \eqref{eq:GS-trick}, we obtain the estimate of the gradient \eqref{eq:bayes-update} as
\begin{multline} \label{eq:bayes-RP-grad-estimator}
    \grad_{\bmmu} \hat{\set{L}}(\bmmu) \triangleq \frac{1-\bmw^2}{\tau \big( 1 - \text{tanh}^2(\bmw^{\text{r}}) \big)} \\
    \odot \frac{1}{|\set{B}|}  \smashoperator[r]{\sum_{(\bmx_{\leq T}, \bmr_{\leq T}) \in \set{B}}} ~~ \grad_{\bmw^\text{r}} \ell \big( \bmf_{\bmw^\text{r}}(\bmx_{\leq T}, \bmr_{\leq T}\big) \big|_{\bmw^\text{r} = \bmw}.
\end{multline}
This estimate is unbiased when $\tau \rightarrow 0$. A comparison of the ST-BiSNN and Bayes-BiSNN rules can be found in Table~\ref{tbl:bisnn-grad}.

\section{Experiments}
\label{sec:exp}
We now validate the performance of the proposed schemes in a variety of experiments, using both synthetic and real neuromorphic datasets. To this end, we implement a state-of-the-art surrogate gradient model, DECOLLE \cite{kaiser2020decolle}, in order to compute the error signal in \eqref{eq:der-general}. DECOLLE connects neurons in the read-out layer with a fixed and random linear auxiliary layer for regression and with a softmax layer for classification. In a similar manner to \cite{rastegari2016xnor}, we scale down the outputs of each layer by a factor $\kappa$ that is selected as $\kappa = 1 / \sqrt{|\set{P}_i|}$ for linear layers, and $\kappa = 1 / \sqrt{C_{\text{in}} k_0\cdot k_1}$ for convolutional layers, with $C_{\text{in}}$ the number of input channels and $\{k_0, k_1\}$ the convolutional kernel sizes in height and width. 

As in \cite{meng2020bayesbinn}, we consider two predictors; the \textit{MAP} predictor obtained for the fixed weight selection $\bmw = \text{sign}(2 \sigma(2\bmw^{r}) - 1)$, which minimizes the variational posterior \eqref{eq:exp-bernoulli}; and the \textit{ensemble} predictor obtained by averaging predictions over $10$ random realizations of the binary weights $\bmw \sim q_{\bmw^r}(\bmw)$.
Code will be made available at \url{https://github.com/kclip}.

\begin{figure}[t!]
\centering
\includegraphics[width=1.\columnwidth]{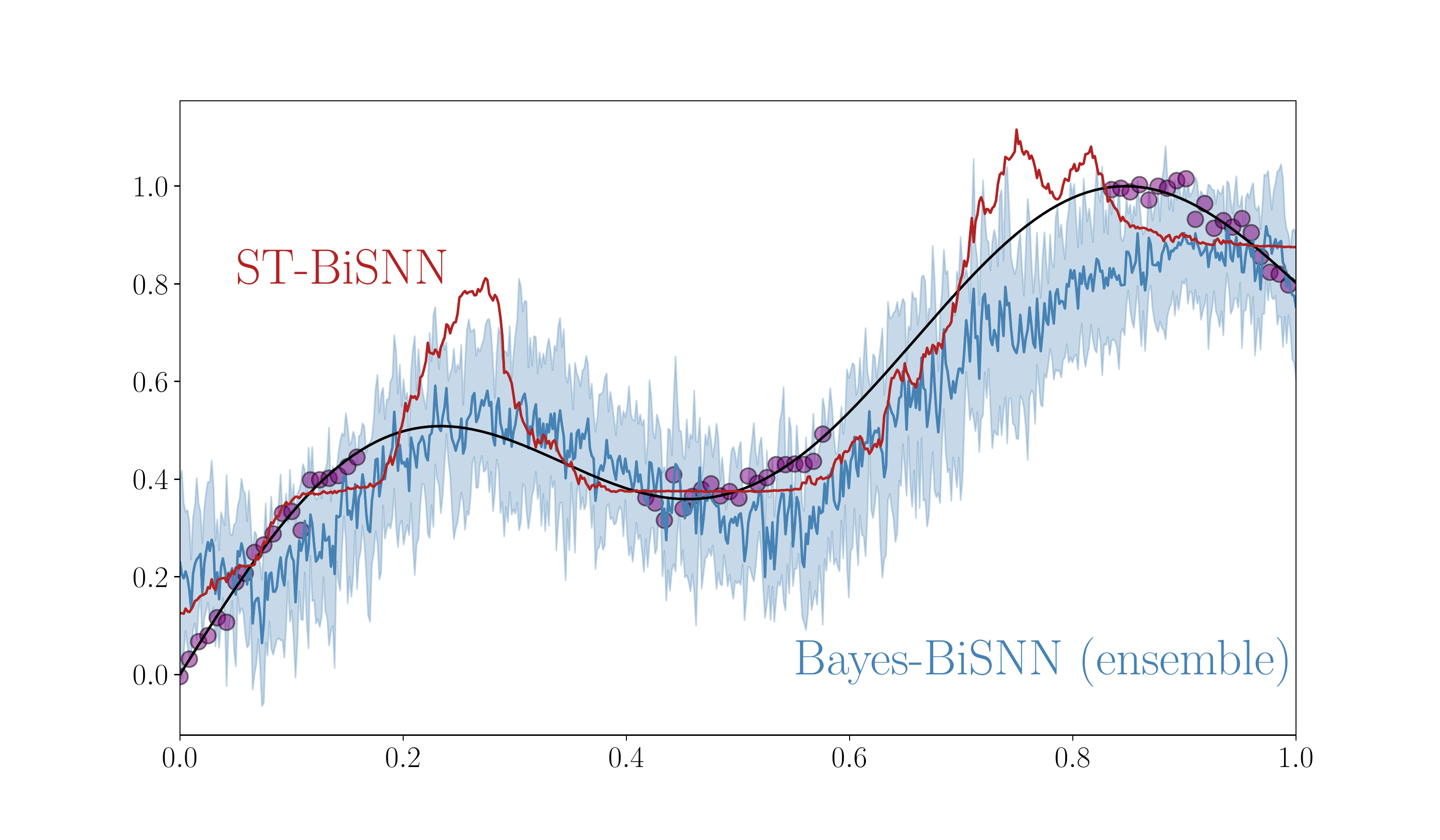}
\vspace{-0.8cm}
\caption{Prediction obtained with ST-BiSNN and with Bayes-BiSNN on $1$-dimensional data. Scatter points in purple represent training data, while the full line in black represents test data. The shaded area represents the standard deviation for the predictions returned by Bayes-BiSNN when the weights are randomly selected from the variational posterior.
}
\label{fig:1d_exp}
\vspace{-0.1cm}
\end{figure}

\begin{figure*}
\centering
\includegraphics[width=1.2\textwidth]{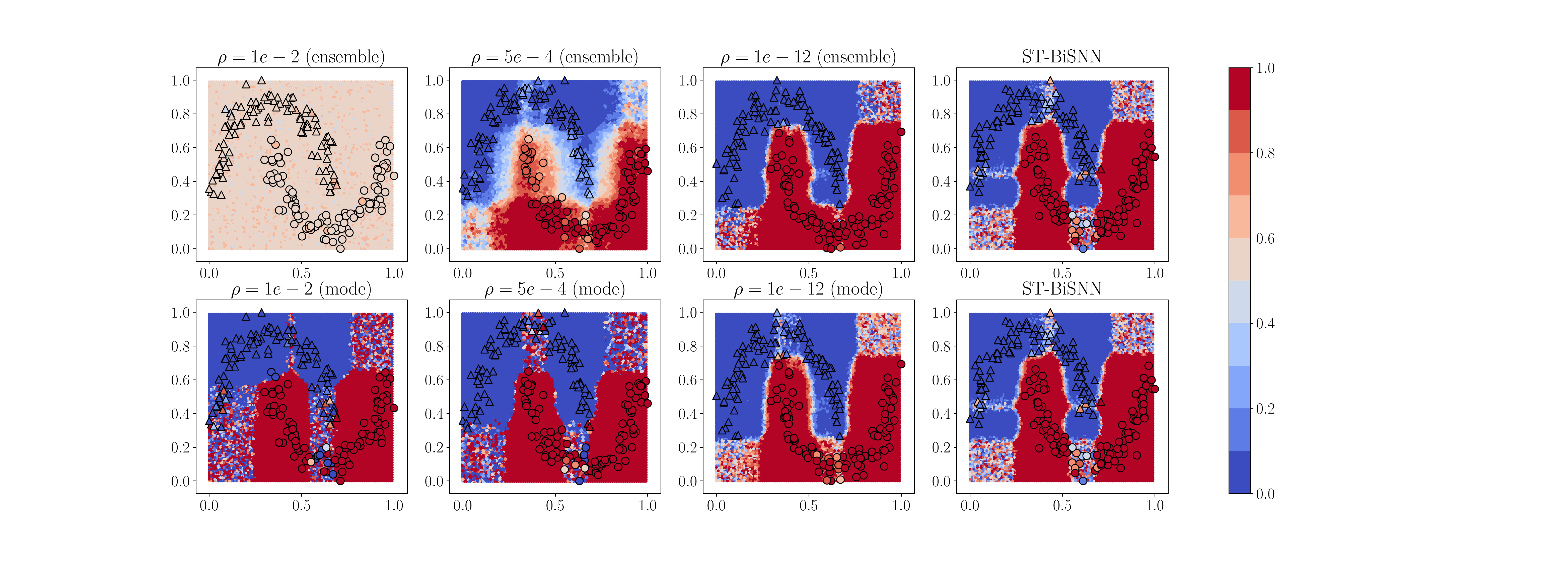} 
\vspace{-1cm}
\caption{Comparison of training with different values of the temperature $\rho$ for Bayes-BiSNN, and with ST-BiSNN.}
\label{fig:twomoons_exp}
\vspace{-0.3cm}
\end{figure*}

\smallskip
\noindent \textit{1D synthetic data:} The SNN consists of two fully connected layers, with each layer comprising $256$ neurons. First, we consider the 1D regression task studied in \cite{daxberger2020expressive-bayes-dl}, where the training data includes three separated clusters of input data points, as illustrated in Fig~\ref{fig:1d_exp}.
% allowing us to evaluate uncertainty outside of them. The outputs are the values of the function $f(x) = x - 0.1 x^2 + \cos(\pi x / 2)$ with training points defined as $x \in [-1, 0] \cup [1.5, 2.5] \cup [4, 5]$, and test data spanning the range $x \in [-1, 5]$. Gaussian noise with standard deviation $0.1$ is added to outputs in the training dataset, and both inputs and outputs are re-scaled in the range $[0, 1]$. 
Input data is converted into a binary spiking signal via population coding \cite{eliasmith2004-neuraleng}, with each scalar input value being encoded over $T=100$ time-steps and using $20$ neurons, while we use the real-valued outputs as targets of the fixed auxiliary layer. In Fig.~\ref{fig:1d_exp}, we compare results for ST-BiSNN and Bayes-BiSNN after $10,000$ epochs of training. Bayes-BiSNN is seen to yield predictions that are more robust to overfitting by capturing the epistemic uncertainty caused by the availability of limited data.

\noindent \textit{2D synthetic data:} Next, we consider the 2D binary classification task on the two moons dataset \cite{two-moons}. Training is done on $200$ samples per class with added noise with standard deviation $0.1$ for $5,000$ epochs. The inputs are again obtained via population encoding over $T=100$ time-steps and via $10$ neurons, while predictions are now obtained via an auxiliary softmax layer. In Fig.~\ref{fig:twomoons_exp}, we compare the results obtained for Bayes-BiSNN with different values of the temperature parameter $\rho$ in \eqref{eq:bayes-opt} and for ST-BiSNN. Triangles indicate training points for class ``$0$'', while circles indicate training points for class ``$1$''. The color intensity highlights the certainty of the network's prediction: the more intense the color, the higher the prediction confidence determined by the softmax layer. We note that the temperature parameter has an important role in preventing overfitting and underfitting of the training data: When $\rho$ is too large, the model cannot fit the data correctly, resulting in inaccurate predictions; while, when $\rho$ is too small, the training data is fit too tightly, leading to a poor representation of the prediction uncertainty outside the training set. A well-chosen value of $\rho$ strikes the best trade-off between faithfully fitting the training data and allowing for uncertainty quantification. It is also seen that the ensemble predictor obtains better calibrated predictions as compared to the MAP predictor. Finally, ST-BiSNN yields similar results to Bayes-BiSNN with low temperature. 

\noindent \textit{Real-world data}: Finally, we consider the neuromorphic datase MNIST-DVS \cite{serrano2015poker}. In Table \ref{tab:comparison-acc}, we compare the test accuracy for DECOLLE trained with full-precision weights using standard frequentist learning, Bayes-BiSNN, and ST-BiSNN for $500$ epochs using the convolutional architecture presented in \cite{kaiser2020decolle}. As can be seen, ST-BiSNN and Bayes-BiSNN maintain competitive accuracy as compared to the network with full-precision weights. Furthermore, the accuracy of Bayes-BiSNN is close to that of ST-BiSNN, with the added advantage, illustrated above, of producing better calibrated decisions.

%Note that the hyperparameters used in this experiment were chosen after a quick search but may not be optimal, leaving the possibility for results to be further improved. In particular, it is unclear whether the chosen per-layer scaling down factor $\delta$ is optimal. Following different rules for the selection of this factor \cite{rastegari2016xnor} may improve performance.

\begin{table}[t]
\caption{Test accuracy of Bayes-BiSNN, ST-BiSNN, and DECOLLE on MNIST-DVS.}
\label{tab:comparison-acc}
%\vskip 0.15in
\begin{center}
\begin{small}
\begin{sc}
\begin{tabular}{lccr}
\toprule
Dataset & Model & Accuracy \\
\midrule
  & DECOLLE (Full) &  $98.90$\% \\
MNIST-DVS & Bayes-BiSNN (Binary) &  $95.10$\%  \\
  & ST-BiSNN (Binary) & $96.00$\% \\
% \midrule
%   & DECOLLE (Full) &  $90.97$\%  \\
% DvsGesture  & Bayes-BiSNN (Binary) &  $68.75$\%  \\
%   & ST-BiSNN (Binary) & $77.08$\% \\
\bottomrule
\end{tabular}
\end{sc}
\end{small}
\end{center}
\vspace{-0.5cm}
\end{table}

\section{Conclusions}
\label{sec:conclusion}
In this paper, we have introduced two learning rules for SNNs that combine the benefits of binary sparse (zero-one) activations and of binary (bipolar) weights for low-power and low-memory supervised learning. In particular, by leveraging Bayesian principles, we have demonstrated the capacity of the proposed model to account for epistemic uncertainty, while maintaining competitive performance as compared to models with full-precision weights. Future work may include further study of the generalization capabilities of the derived rule on larger datasets.

\vfill \pagebreak

% BiBTeX files (here: strings, refs, manuals). The IEEE bib.bst bibliography

%\newpage
%{ \small
\bibliographystyle{IEEEbib}
\bibliography{ref}
%}

%\newpage
%\appendix
%\input{etc}

\end{document}